# Whisper Finetuning on Nepali Language

Sanjay Rijal *[†1], Shital Adhikari *[2], Manish Dahal *[3], Manish Awale[4], and Vaghawan Ojha [†5]

[1]sanjay.rijal@ekbana.info, rijalsanjay42@gmail.com
[2]shital.adhikari@ekbana.info
[3]manish.dahal@ekbana.info
[4]manish.awale@ekbana.info
[5]vaghawan.ojha@ekbana.net
E.K. Solutions Pvt. Ltd., Lalitpur, Nepal

## Abstract

Despite the growing advancements in Automatic Speech Recognition (ASR) models, the development of robust models for underrepresented languages, such as Nepali, remains a challenge. This research focuses on making an exhaustive and generalized dataset followed by fine-tuning OpenAI's Whisper models of different sizes to improve transcription (speech-to-text) accuracy for the Nepali language. We leverage publicly available ASR datasets and self-recorded custom datasets with a diverse range of accents, dialects, and speaking styles further enriched through augmentation. Our experimental results demonstrate that fine-tuning Whisper models on our curated custom dataset substantially reduces the Word Error Rate (WER) across all model sizes attributed to larger data variations in terms of speaker's age, gender, and sentiment, acoustic environment, dialect, denser audio segments (15-30 seconds) that are more compatible with Whisper's input, and manual curation of audios and transcriptions. Notably, our approach outperforms Whisper's baseline models trained on Fleur's dataset, achieving WER reductions of up to 36.2% on the small and 23.8% on medium models. Furthermore, we show that data augmentation plays a significant role in enhancing model robustness. Our approach underlines the importance of dataset quality, variation, and augmentation in the adaptation of state-of-the-art models to underrepresented languages for developing accurate ASR systems.

## 1 Introduction

Automatic Speech Recognition (ASR) systems have experienced remarkable advancements in recent years, driven largely by the development of large-scale, pre-trained models such as OpenAI's Whisper [1]. These models, trained on extensive multilingual datasets, have demonstrated impressive performance across a wide range of languages, supporting applications in voice assistants, automated transcription, and accessibility tools for the hearing impaired. However, many low-resource languages, including Nepali, Hindi, and Albanian, continue to face challenges in achieving high transcription accuracy. These challenges stem from the limited availability of high-quality training data and the linguistic complexities of these languages, such as their rich morphology, multiple dialects, and unique phonetic structures.

---

*These authors contributed equally to this work.
†Corresponding authors

Prior works, including Whisper's original implementation, have trained models using datasets like Fleurs [12]. Fleurs provide only 10.38 hours of Nepali speech data with relatively short audio segments (2 to 10 seconds) and minimal variation in acoustic environments. As a result, the Word Error Rates (WER) for Nepali transcriptions in these models remain significantly higher compared to widely spoken languages. The ASR systems pre-trained in a supervised way across many datasets or domains are robust and can better generalize than models trained on a single source [2, 4, 3]. This is possible by combining the high-quality datasets as much as possible. OpenAI's Whisper [1] tried mitigating this by scaling weakly supervised speech recognition to the order of 680,000 hours of labeled data covering more than 96 languages. However, for low-resourced languages such as Nepali language, language-specific finetuning incorporating sentiments, accents, pronunciation, acoustic environment, gender, and age proves to perform better, especially with long audio segments [5, 6, 7].

In recent years, several studies [5, 8, 9] have focused on fine-tuning ASR models for low-resource languages, yielding significant improvements in transcription accuracy. Notable works include the Gram Vaani [5] and Vistaar [8] projects, which focused on languages like Hindi, Marathi, and Gujarati. These studies have demonstrated the efficacy of using domain-specific datasets and fine-tuning techniques to reduce WER. Gram Vaani, a social enterprise in rural India providing voice-based interactions for call center automation, organized an ASR challenge in 2022 to improve speech recognition for agricultural and healthcare advisory systems. The study employed both traditional time-delay neural network-hidden Markov models (TDNN-HMM) and fully neural end-to-end (E2E) models, showing remarkable improvements in WER, between 30.1% to 37.3%, across different models. Similarly, Patel et al. [11] showed a better performance of 30.3% WER using the E2E Conformer model, surpassing the baseline of 34.8% set during the Gram Vaani challenge. Vistaar [8] also provides benchmark datasets for 59 Indian languages, facilitating comparative studies in diverse acoustic and linguistic environments. These initiatives emphasize the importance of creating rich, domain-specific datasets that reflect the linguistic diversity of the target languages. Moreover, the application of advanced models like Whisper [1] and wav2vec [10] in these studies has underscored the value of large-scale pre-training followed by language-specific fine-tuning.

Building on these advancements, our work focuses on fine-tuning Whisper models for Nepali ASR which can also be implemented in other low-resourced languages. We leverage a diverse and extensive dataset that includes publicly available speech corpora such as Google Fleurs [12], Mozilla Common Voice [13], and OpenSLR [14, 15], along with a custom dataset built from self-recorded audios. The custom corpus encompasses a wide variety of audio environments, speaker demographics, and speech styles, significantly expanding the diversity and volume of data available for training. The fine-tuned `small` model on the custom dataset shows a significant improvement of 68.5% compared to other ASR datasets. Moreover, we implement data augmentation techniques, to further enhance the model's robustness. By fine-tuning the Whisper models on this curated dataset, we significantly reduce WER across multiple model sizes including `tiny` (68.5%), `base` (70.2%), `small` (36.2%), and `medium` (23.8%). Our approach demonstrates the critical role that dataset quality, variation, and augmentation play in improving ASR performance for underrepresented languages.

# 2 Dataset

The dataset used in this work consists of publicly available ASR datasets such as Google Fleurs [12], Mozilla Common Voice [13], and OpenSLR datasets SLR43 [14] and SLR143 [15]. Additionally, we also prepared a custom dataset from a diverse and extensive pool of self-recordings and publicly available sources, representing a wide range of speakers and topic. As shown in Table 1, the cumulative dataset contains a total of 33.97 hours of raw audio, distributed across:

Table 1: Raw audio data lengths on different datasets.

| Dataset | Size (Hrs) |
|---|---|
| Fleurs [12] | 10.38 |
| Common Voice [13] | 1.28 |
| SLR43 [14] | 2.82 |
| SLR143 [15] | 1.25 |
| Custom Dataset | 18.24 |

The data includes speech from various environments, ranging from clean to noisy conditions, ensuring that the transcriptions are accurate despite the background noise. This diverse dataset is critical for training robust speech recognition models that generalize well across different acoustic settings.

## 2.1 Custom Dataset Preparation

The custom dataset primarily consists of read speech and lecture-style recordings, sourced from publicly available sources and self-recorded audio. The self-recorded portion of the dataset includes readings from online news articles, academic texts, and other sources. The dataset is diverse in terms of speaker demographics, containing data from various age groups, backgrounds, sentiments, and genders, as detailed in Table 2. Moreover, as shown in Figure 1, our custom dataset introduces significantly larger vocabularies compared to other open-source datasets used here, allowing the model to generalize across a broader range of speech. We discuss this improved generalization in detail in subsection 4.1.

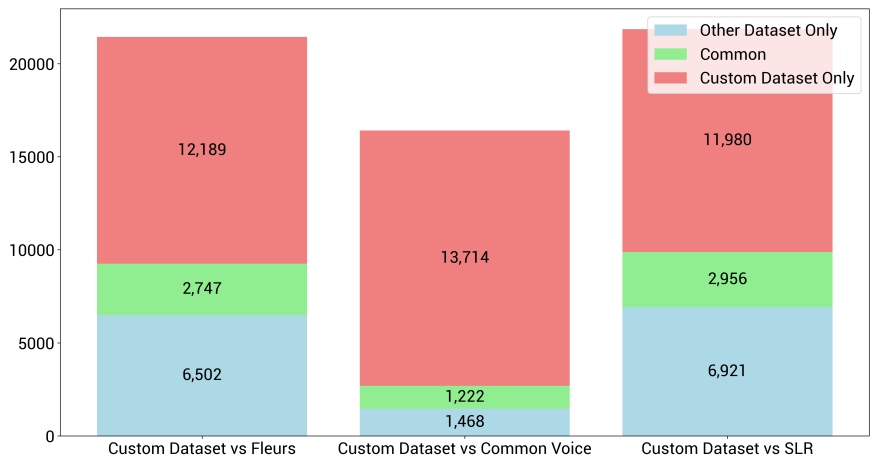

Figure 1: Distribution of unique words on custom dataset compared against other open-source datasets. Here unique words don't include articles, conjunctions, prepositions, and exclamations.

Table 2: Metadata of custom dataset

| Metadata | Ranges |
| --- | --- |
| Gender | Male, Female, Unknown |
| Speaker Age | 25-60 |
| Background | Clean, White Noise, Crowded |
| Sentiment | Happy, Sad, Normal, Angry |

Before merging the custom dataset's raw audio to the corpus, we perform a few preprocessing to ensure consistency in data and metadata and removal of silence and unrecognized audio segments. We use Audacity[17], an audio editing tool to segment audio into chunks of 30 seconds and remove artifacts like silence, and unrecognized audio segments. We first remove the silences > 1 second and then manually remove the unrecognized or corrupted audio segments. Transcriptions were either sourced from publicly available documents of the corresponding audios or generated manually where required. For documents that contain transcription errors, we apply manual corrections to improve the alignment with the audio, ensuring high-quality training data. After this audio pre-processing, the size of the dataset was reduced to 13.58 hours.

To increase the variability and robustness of the dataset, we employ data augmentation techniques using `torchaudio` [16]. Specifically, we added 8000Hz white noise to the segmented audio, increasing the data volume while introducing minor distortions that mimic real-world noisy environments. This augmentation process expanded the total size of the custom dataset to 27.17 hours, and when combined with the other datasets, the total size of our corpus reached 42.9 hours.

## 2.2 Train and Evaluation Dataset

We conducted a series of experiments for each dataset, fine-tuning Whisper's pre-trained models across multiple configurations. The training data was split into 80% for training and 20% for evaluation. The same data partitioning strategy was applied across all datasets to ensure a fair comparison between our models and Whisper's baseline models, `tiny`, `base`, `small`, and `medium`. In addition, we fine-tune the above models on combined datasets: a) Fleurs, and Common Voice, b) Fleurs, Common Voice, and SLR (SLR43 + SLR143) c) Fleurs, Common Voice, SLR, and custom i.e. `all_combined`, d) `augmented` i.e. `all_combined` and its augmentation.

We shuffle the training and evaluation datasets separately to reduce the correlation between them, which minimizes the risk of overfitting and allows the model to generalize better to unseen data. Moreover, for a fair and better generalization, we use evaluation data as 30% the size of the shuffled training dataset for individual corpus. This is important because if we split the evaluation data only from a specific dataset then it won't generalize with the same WER accuracy on other datasets. For example, in the case of Fleurs if both training and evaluation are samples of the Fleurs dataset, then it will perform poorly on the rest of the dataset, especially when the audio segments are more than 10 seconds.

As shown in Figures 3 and 5, the average WER of the training data decreases progressively as the model learns to handle the diverse linguistic characteristics present in the Nepali language, while the validation data consistently demonstrate improved performance as well.

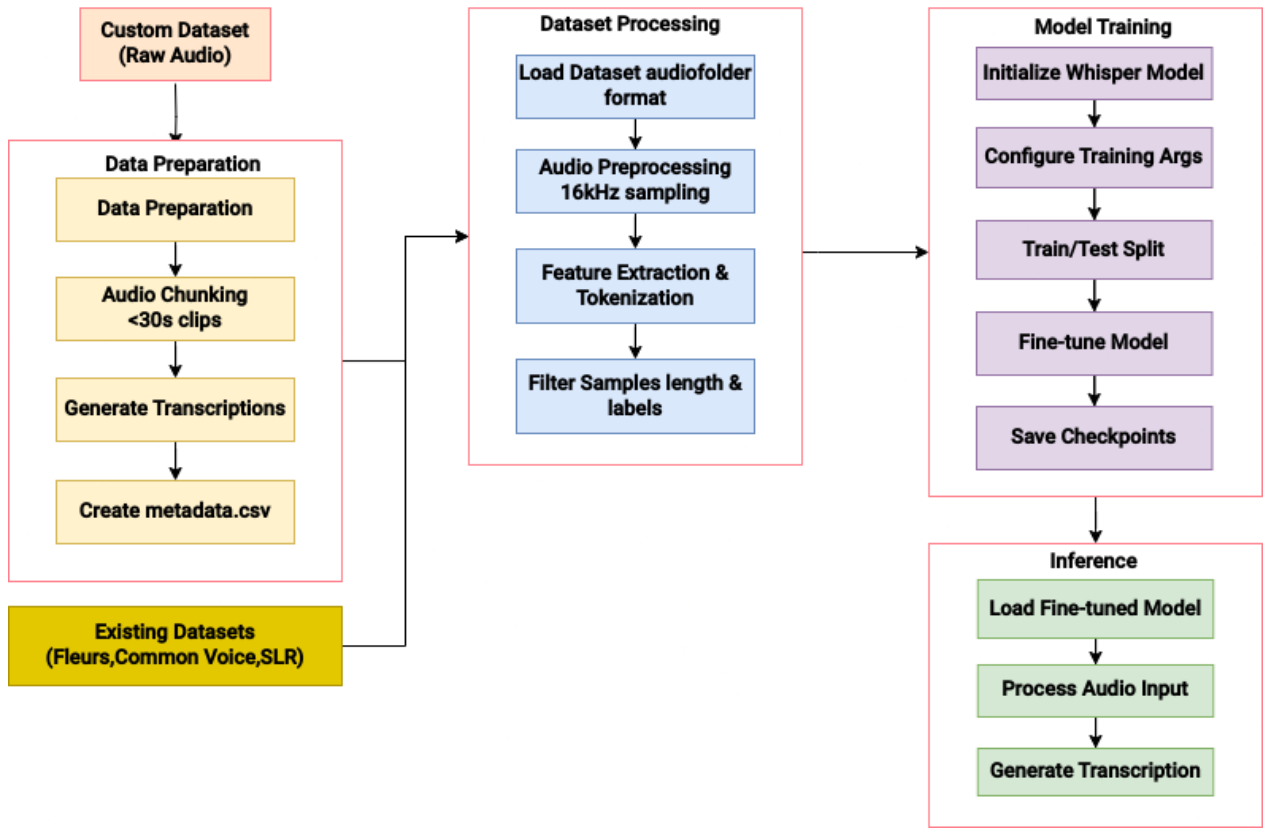

Figure 2: Fine-tuning pipeline

## 3 Fine-tuning Pipeline

Our overall pipeline is shown in Figure 2. The pipeline consists of four main stages: data preparation, dataset processing, model training, and inference. As explained in subsection 2.1, our custom dataset requires a few audio pre-processing which includes audio chunking, silence removal, and unrecognized audio and transcriptions filtering. After forming the custom dataset we merge it with other open-source datasets and all the transcripts into a single `metadata.csv` file.

All datasets are loaded in `audiofolder` format, followed by preprocessing to resample audio clips to a uniform 16kHz frequency. The pipeline then performs feature extraction and tokenization to prepare the data for model training. The samples are then filtered by length and label suitability to ensure consistency across the dataset.

For model training, we employ the Whisper architecture [1] by fully training its base model rather than using a pre-trained model. The Whisper model is initialized with specific training configurations, and a train-test split is defined as described in subsection 2.2. During training, the pipeline utilizes Whisper's architecture, loss functions, and optimizers enhancing the convergence. Checkpoints are then saved periodically to allow resumption and model improvement tracking. Finally, the model is used for inference with an audio input to generate transcriptions.

## 4 Experiments

We evaluate our fine-tuned Whisper models on various datasets, both independently and in combination (`all_combined`). To assess the impact of data augmentation, we also compare performance metrics, specifically loss and word error rate (WER) as primary metrics, before and after augmentation on `all_combined`. Additionally, we perform a comparative analysis with the Whisper models presented in OpenAI's original paper, which used the Fleurs dataset [12] for the Nepali language.

All the experiments are performed on an Intel i9-10900 CPU @ 3.70GHz paired with 64GB DDR4 RAM and a 24GB NVIDIA GeForce RTX 3090 @ 33MHz GPU.

## 4.1 Comparison on Different Datasets

We individually compare the results of fine-tuned `small` models across different individual and combined corpora. We initially fine-tuned the models on individual datasets as outlined in Table 1. However, given the relatively small size of these datasets, some models displayed signs of overfitting after a certain number of epochs, despite adjustments to hyperparameters, leading to suboptimal results. The overfitting can be attributed to the limited diversity and volume of the Nepali language datasets shown in Table 1, which proved insufficient for robust deep-learning training. To mitigate this issue, we earlystop the training process at 1500 epoch and the results are shown in Figure 3. Following early stopping, all datasets show a decreasing trend in loss and WER as shown in Figures 3 and 4, with the custom dataset demonstrating a marked improvement in WER across all the models compared to the other datasets. This enhancement is due to the custom dataset's greater comprehensiveness, encompassing a wider range of metadata and a larger vocabulary.

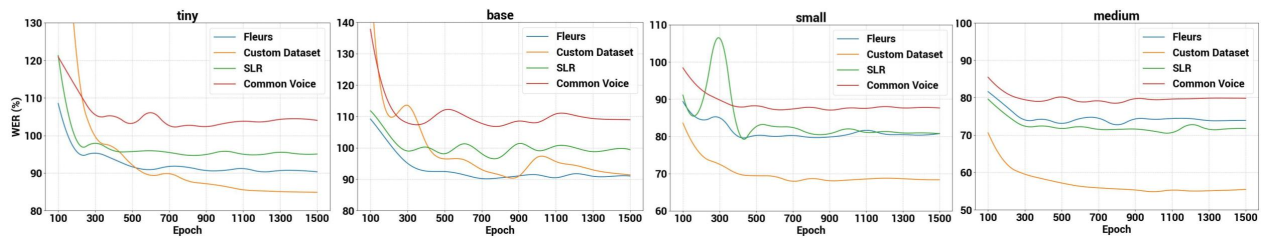

Figure 3: Comparison of WER on different individual datasets fine-tuned on `tiny`, `base`, `small`, and `medium` models. WER on our custom dataset is lower than other datasets across all the models, and more significant on larger models.

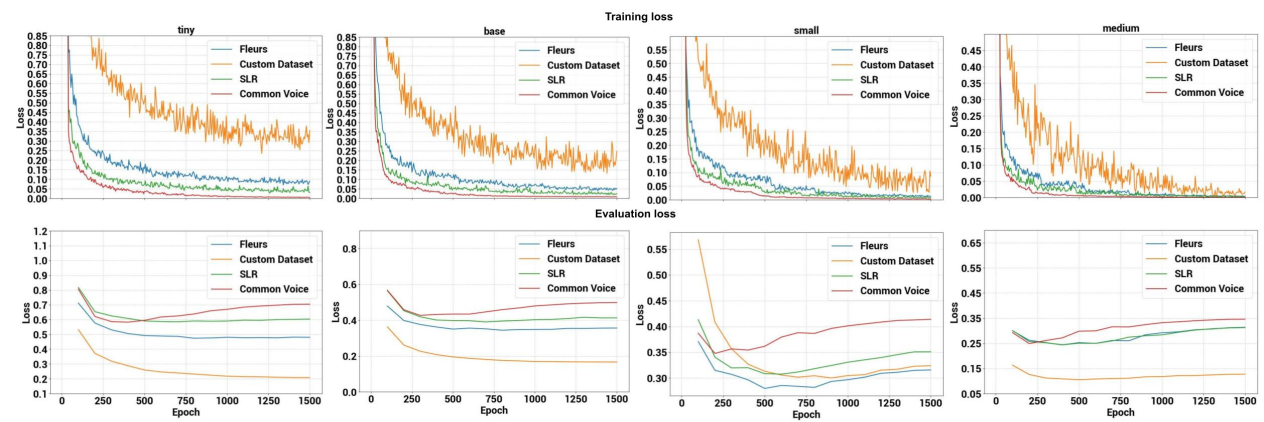

Figure 4: Comparison of training and evaluation loss on individual datasets fine-tuned on `tiny`, `base`, `small`, and `medium` models.

Considering the overfitting issue on limited individual datasets, we combine these datasets on a cumulative basis into a single corpus as explained in section 2.2 and perform a comparative analysis. By training the models on this combined dataset for 4000 epochs, we achieve better generalization without encountering overfitting, as shown by the significant drop in WER in Figure 6.

Table 3: Transcription prediction comparison on `small` fine-tuned model across different datasets. Ground Truth refers to the original transcriptions from the dataset used for benchmarking.

| Datasets | Predictions |
| --- | --- |
| Ground Truth | देशभर पश्चिमी वायु र स्थानीय वायुको आंशिक प्रभाव रहेको छ। मौसम पूर्वानुमान महाशाखाका अनुसार सोमबार देशका पहाडी भूभागमा आंशिक बदली रही बाँकी भूभागमा आंशिक बदलीदेखि मुख्यतः सफा रहने छ। कोशी, बागमती, गण्डकी, कर्णाली र सुदूरपश्चिम प्रदेशका पहाडी भूभागका एक–दुई स्थानमा हल्का वर्षाको साथै उच्च पहाडी तथा हिमाली भूभागका एक–दुई स्थानमा हल्का हिमपातको पनि सम्भावना रहेको छ। |
| Fleurs | देशबर पश्चिमी वायु र इस्थानी वाईको आम्सिक प्रवाव रहेको छ मौसोम पूरुवानुमा शाखाकानुसार स्रोमबार देशको पाडी भूबागमा आम्सिक बदलिरै बाँकी भूबागमा आम्सिक बदलिदेखि मौसोमुख खेतया सफार अनिछ कोशीबागमती गण्डकी खर्णाली र सुधुपशिम प्रदेशका पाडी भूबागका एक दुईस्तानमा हल्का वर्षाको साथै उचपाडी तथाइ हिमाली भूबागका एक दुईस्तानमा हल्का हिमपादको पनि सम्भावना रहेको छ। |
| Common Voice | देशबर पश्चिमी भाङु रेस्थानी भाइको आम्सिक प्रभाब रहेको छ मौसम पुरुभानुमानमा साखाकानुसार शुहुम्भार देशको पाडी भोबागमा आम्सिक बढाली रै वाकी भोबागमा अःम्सिक बढाली देखि मौसम मुख्खेठ्या सफार निछ कोशीबागमतिग गण्डुकी खणाली र सुदुपशिम प्रद्यचका पाडी भोबागका एक्दौस्थानमा हल्का वष्र्वाको साथै उचपाडी तथाइ हिमाली भोबागका एक्दुस्थानमा हल्का हिम्पादको पनि सम्भाबना रहेको छ। |
| SLR | देशबर पश्चिमी बायु रेष्ठानी बाइको आम्सिकप्रबाव रहेको छ मौसम पूरुबानुमा शाखाकानुसार स्रोमबार देशको पाडी भूबागमा आम्सिक बदलिरै बाँकी भूबागमा आम्सिक बदलिदेखि मौसम मुख्खेतया सभार निछछकोशीबागम्ति गण्कि खणालिर सुदुपशिम प्रदेशका पाडी भूबागका एक्दुस्थानमा हल्का वर्षाको साथै उचपाडी तथाइहिमाली भूबागका एक्दुस्थानमा हल्का हिम्पादको पनि सम्बाभना रहेको छ। |
| Custom Dataset | देशबर प्रष्िमी भागीय र स्थानीबाँको आंसिक प्रभाव रहेको छ मौसम पूर्वनुमा आशाखाका अनुसार स्रोम्बार देशको पारी भूभागमा आंसिक बदलिरै बाँकी भूभागमा आंसिक बदलिदेखि मौसम्मुखखेथ्या सफारा अनि छ कोसी बागमती गण्ढकी खर्णा लिर सुदुपस्थीम प्रदेशका पारी भूभागका एक दुई स्थानमा हल्का वर्षाको साथै उच पारी तथा हिमाली भूभागका एक दुई स्थानमा हल्का हिम्पादको पनि सम्भावना रहेको छ । |
| Fleurs+Common Voice | देशबर पश्िमी भाग्यु र इस्तानी भागीको आम्सिक प्रभाव रहेको छ मौसम पूरुबानुमा औंमा शाखाकानुसार श्रोमबार देशको पाडी भौभागमा आम्सिक बदलिरै बाँकी भौभागमा आम्सिक बदलिदेखि मौसम्मुख खेथया सफार हनि छ कोशी बाग्मति गण्ढकी खणाली र सुदुपशिम प्रदेशका पाडी भौभागका एक्दुस्तानमा हल्का वर्षाको साथै उचपाडी तथाथै हिमाली भौभागका एक्दुस्तानमा हल्का हिम्पादको पनि सम्पार रहेको छ। |
| Fleurs+Common Voice+SLR | देशबर पश्चिमी बायु र इस्तानी बाईको आम्सिक प्रभाव रहेको छ मौसम पूरुवानुमा नमा शाखाकाअनुसार स्रोमबार देशको पाडी भूभागमा आम्सिक बदलिरै बाँकी भूभागमा आम्सिक बदलिदेखि मौसम मुख्यतया सफार निछ कोशी भागमति गण्ढकी खणाली र सुदुपशिम प्रदेशका पाडी भूभागका एक दुईस्तानमा हल्का वर्षाको साथै उचपाडी तथाइ हिमाली भूभागका एक दुईस्तानमा हल्का हिम्पादको पनि सम्भावना रहेको छ। |
| all_combined | देशबर पश्चिमी बाहिँ र स्थानी बाहीको आम्सिक प्रभाव रहेको छ मौसम पूर्वानुमानमा आशाखाका अनुसार स्रोमबार देशको पारी भूभागमा आम्सिक बदलिरही बाँकी भूभागमा आम्सिक बदलिदेखि मौसम भूखेथया सफार अनि छ कोसी बागमाती गण्ढकी खर्णाली र सुदुपशिम प्रदेशका पारी भूभागका एक दुई स्थानमा हल्का वर्षाको साथै उच पारी तथाई हिमाली भूबागका एक दुई स्थानमा हल्का हिम्पादको पनि सम्भावना रहेको छ । |

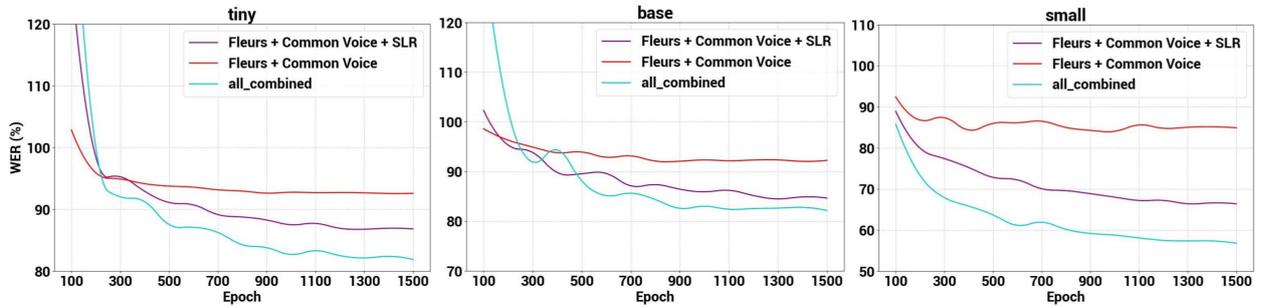

Figure 5: Comparison of WER on cumulative datasets with Fleurs as the baseline. `all_combined` outperforms others which becomes more significant as the model complexity increases.

As illustrated in Figures 3 and 5, the custom dataset's effectiveness in reducing WER is comparable to the cumulative impact of the Fleurs, Common Voice, and SLR datasets, underscoring the dataset's exhaustiveness and efficiency. Furthermore, adding data from Fleurs, initially used by Whisper, resulted in a progressively smoother WER trend. Notably, the comprehensive combination of all datasets, represented as `all_combined`, achieved a significant WER reduction of approximately 56%. To further assess prediction accuracy across individual and combined corpora, we present a comparative table, Table 3 comparing the predictions from each dataset with the ground truth. This table underscores the custom dataset's prediction accuracy and demonstrates the datasets' combined effectiveness, as reflected in the corresponding WER plots. Additionally, as seen in Table 3, a comparison of models trained on open-source datasets versus the custom dataset shows that models trained on open-source datasets struggle with audio inputs exceeding 15 seconds, producing random predictions beyond a certain audio length. In contrast, our custom dataset better aligns with Whisper's input requirements, resulting in improved and more accurate predictions on longer audio segments.

Table 4: WER comparison on models fine-tuned with different datasets for 1500 epochs.

| Datasets | Models | | | |
|---|---|---|---|---|
| | tiny | base | small | medium |
| Fleurs | 90.34 | 91.01 | 80.7 | 73.92 |
| Common Voice | 104.04 | 108.92 | 87.9 | 79.85 |
| SLR | 95.07 | 99.38 | 81.0 | 71.77 |
| Custom Dataset | 84.89 | 91.37 | 68.5 | 55.47 |
| Fleurs+Common Voice | 92.58 | 92.24 | 85.0 | – |
| Fleurs+Common Voice+SLR | 86.82 | 84.65 | 67.0 | – |

## 4.2 Augmentation Results

Following the combination of datasets, we applied data augmentation techniques to further enhance model performance. Specifically, we introduce an 8000Hz white noise to the raw audio using `torchaudio` [16]. This augmentation is done on `all_combined` dataset only for the audios whose resampled noise lengths are less than the original audio length. Although very simple, introducing white noise to the raw audio data not only increased the volume of the dataset but also improved the model's performance, as reflected by the decrease in WER in Figure 6. This highlights the role of data augmentation as a critical step in model fine-tuning, especially when working with limited data. While the observed WER reduction is modest ( 4%), we anticipate that more sophisticated augmentation methods could yield more substantial improvements. However, as data augmentation is not the primary focus of our study, we only demonstrate a basic augmentation method to show how even simple techniques can improve model performance.

Given that we have already compared WER and predictions across individual and combined datasets, we restrict our evaluation of augmentation results to the `small` fine-tuned model on the combined dataset (`all_combined`). The observed improvements are consistent across other datasets as well, as evidenced by Figure 7, which illustrates training on the Fleurs dataset with augmentation up to 7,000 epochs without overfitting. Predictions from models trained on `all_combined` and `augmented` datasets are presented in Table 5, demonstrating subtle yet meaningful improvements in prediction accuracy for the model trained with augmented data.

Table 5: Impact of data augmentation on prediction (with `small` fine-tuned model).

| Datasets | Predictions |
| --- | --- |
| Ground Truth | विदेशी मुद्रा सञ्चितिको लाभ उठाउनसमेत सरकार असफल । चरम कुपोषण प्रभावित विश्वका करिब २० लाख बालबालिका मृत्युको जोखिममा रहेको यूनिसेफको चेतावनी उपचारात्मक तयारी भोजन खरिदका लागि आर्थिक सहयोगको आह्वान । |
| all_combined | विदेशी मुद्रा संस्थितिको लाभ उठाउन समेत सरकार असफत जरमको पोषण प्रभावित विश्वका करिब २० लाख बालबालिका ब्रत्युको चोखिमा रहेको युनिस्टेफको चेता वनी उपजारात्मक तयारी भ्रजन खरिदका लागि आर्थिक सहयोग वावान । |
| augmented | विदेशी मुद्रा संस्कृतिको लाभ उठाउन समेत सरकार अवसफत जरमको पोषण प्रभावित विश्वका करिब २० लाख बालबालीका ब्रत्युको जोखिम्बा रहेको युनिस सेफको चेतावनी उपजारात्मक तयारी भ्रजन खरिदका लागि आर्थिक सहयोग वावहान । |
| Ground Truth | र साफ महिला च्याम्पियनसिपको सातौं संस्करण भोलिदेखि काठमाडौंमा उद्घाटन खेलमा भारत र पाकिस्तानबीच प्रतिस्पर्धा बलियो टोली नेपाललाई इतिहास रच्ने मौका । |
| all_combined | र साथ महिला च्याम्पियन्सिपको सातौँ संस्करण भोलिदेखि कार्ख्मडौंमा उद्घाटन खेलमा भारत र पाकिस्तान बिच प्रतिस्पर्दा बलियो टोली नेपाललाई इतिहास रस नेमौका । |
| augmented | र साथ महिला च्याम्पियन्सिपको सातौँ संस्करण भोलिदेखि कार्टमाडौंमा उद्घाटन खेलमा भारत र पाकिस्तान बिच प्रतिस्पर्दा बलियो टोली नेपाललाई इतिहास रस्नेमौका । |

## 4.3 Comparison with Whisper

Alongside our custom dataset, we perform a comparative analysis on the benchmark dataset, Fleurs [12] originally used by Whisper for a comprehensive and fair evaluation. We finetune the `tiny`, `base`, `small`, and `medium` whisper models. However, because of the GPU limitation, we couldn't fully train the `large-v1` and `large-v2` models. Due to an overfitting issue as explained in subsection 4.1, we use augmented datasets for training the models. Table 6 compares WER on various models between the whisper and our approach. We also show a progressive graphical comparison of WERs on different fine-tuned models in Figure 7. Our fine-tuned models show a significantly improved WER on all the trained models for the Nepali language. This improvement is most evident in the `small` and `medium` models, where the WER was reduced from 69.5 to 36.2 and 54.4 to 23.8 respectively, representing a substantial performance gain.

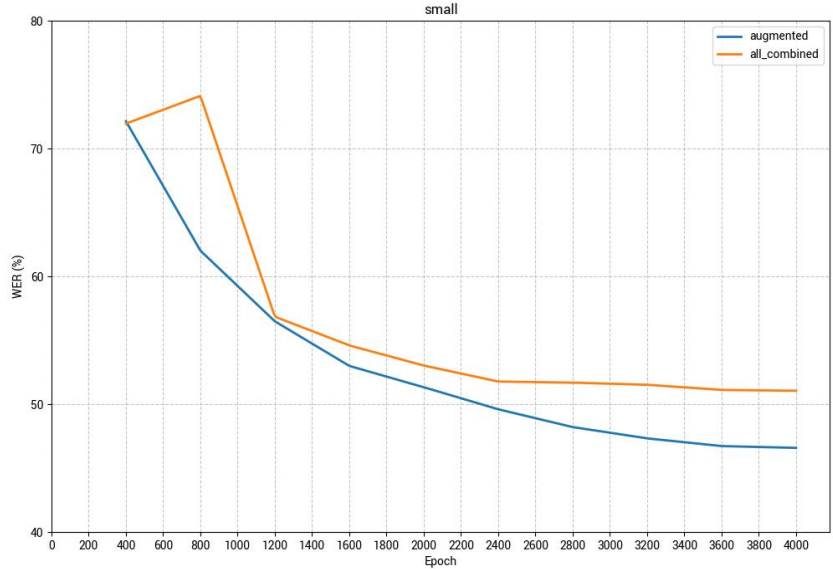

Figure 6: Comparison of WER on `all_combined` and augmented datasets. Here `augmented` refers to `all_combined`+custom augmented.

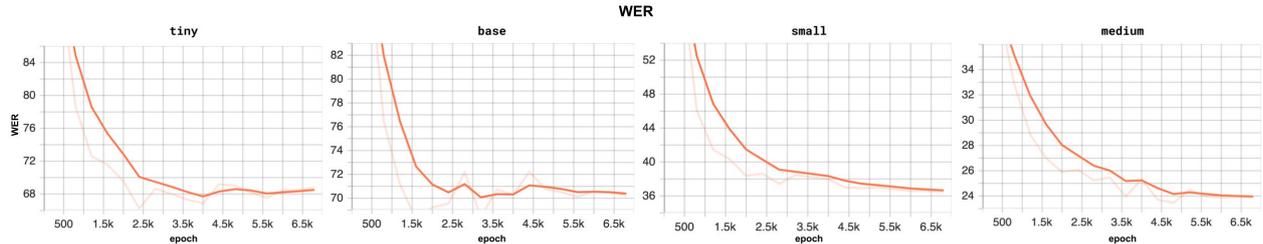

Figure 7: WER on Fleurs dataset [12] fine-tuned on `tiny`, `base`, `small`, and `medium` whisper models.

Table 6: WER comparison between OpenAI's whisper [1] `small` and our fine-tuned models on Fleurs datasets.

| Models | Whisper | Fleurs |
|--------|---------|--------|
| tiny   | 101.8   | **68.5** |
| base   | 102.4   | **70.2** |
| small  | 69.5    | **36.2** |
| medium | 54.4    | **23.8** |

As seen in the prediction comparison Table 7, although the ground truth is in Devanagari script, the transcriptions generated by Whisper's `tiny` and `base` are in Latin script, which explains the WER > 100 in these models. However, our fine-tuned models across all the model sizes generate predictions in Devanagari scripts while still incorporating the numbers and punctuations. As evident from Tables 6 and 7, our finetuned models generate better predictions than Whisper's original models with significantly lower WER.

Table 7: Transcription prediction comparison between OpenAI's Whisper [1] and our fine-tuned models on Fleurs dataset.

| Model | Whisper | Ours (Fleurs) |
|---|---|---|
| Ground Truth | र साफ महिला च्याम्पियनसिपको सातौं संस्करण भोलिदेखि काठमाडौंमा उद्घाटन खेलमा भारत र पाकिस्तानबीच प्रतिस्पर्धा बलियो टोली नेपाललाई इतिहास रच्ने मौका । | |
| tiny | Rassa, you've got to understand, you've got to understand. Rassa, you've got to understand. Rassa, you've got to understand. | र स्वाथ महिला च्यापेन्सिटको सातो सम्सकरण बोली देखिका थमन्डोबा उत्घातन केलमा भार त्रपाकस्ताल विष प्रतिस्पर्दा बलियो टोली नेपालला इतिहास रोफ्ने मौका |
| base | Raasad Mahila Champin Sip ko Saatom Samskarant Boli Dhe Ki Kart Mandoma. Uddhgatant Kheelma Bhaar Atra Pakistan Bis Pratis Pratis Pradha. Balyotoli Neparla Itihas Rox Ne Mau Ka. | र साथ महिला च्याप्प्यान्पेन्सिपको सातौं सम्सकरण बोलीदेखि कात बन्डौमा उत्घातन खेलमा भारत पाकिस्थान विज प्रतिस्तिश बलियो टोली नेपालाई इतिहास र स्रोस्थधै मौखुने मौगका। |
| small | रसाथ महला चामपिन्सिप को साथो सम्सकरन भोली देकी काध्मन्दोमा उद्घाटन केल्मा भारद्र पाकिस्तान भीज्प्रतिस्पर्दा बलियो तोली निपाल लाई इतिहास रोस ने माउका | र साथ महिला च्याम्पियनसिपको सातौं सम्सकरण भोलीदेखि कार्टमण्डोमा उत्गाटन खेलमा भारत र पाकिस्तान बीच प्रतिस्पर्दा बलियोटोली नेपाललाई इतिहास रस्नेमौका। |
| medium | रसाप महिला चाम्पिनसिप को सातो सम्सकरन बोली देखी काट्मनडों मा उध्गातन खेल मा भारत र पाकिस्तान बीज परतिस्पर्दा बल्यो तोली निपारला इतिहास रौस ने मोवका | र साथ महिला च्याम्पियनसिपको सातौं संस्करण बोलीदेखि कार्टमन्डोउमा उद्घातन खेलमा भारत र पाकिस्तान बीच प्रतिस्पर्दा बलियो टोली नेपाललाई इतिहास रोस्ने मौका। |
| Ground Truth | प्रतिकूल मौसमले उडान ठप्प हुँदा मन्थलीमा ३ दिनदेखि अलपत्र पर्यटकलाई सेनाको जहा– जबाट लुक्ला पुर्याईने बिपी राजमार्गमा साँझ ६ देखि बिहान ४ बजेसम्म सवारी चलाउन रोक । | |
| tiny | pratikul mousam le udan thapa huda muntali matindin deki alapatra podyata klasena kudzahazbatun lu klapuryaini dpi razmargama saa jatshaudi ki bihanat saar bodis sama savali saalaun arho | व्रतिकोल मौसम्ले उदान थप्प हुँदा बन् थली मा कि न्धिन देखि अलपत्र परियतकले सेनाको जहासबाट लुकला पूर्याईनी दिपि राजमार्गमा साज छौदेकी विहाल चार वजिससम्ब सबाटि छ लाउन रो। |
| base | Pratikul Mausam le Udan Thappahuda Manthali Matin Dindi ki alapatra Pariyataklaisena Kuzhaha Zbata Nukhla Puriayini Deepi Rajmargama Saadach Saadakhi Bhihana Chhar Bhajis Samasavari Salau Na Roo | प्रतिकूल मौसमले उडान ठप्वपप हुँदा मन्थलीमा तीन दिन दिन देखि अलपत्र परियोटकलाई सेनाको जहाजबाज बाट लुक्का पूर्याईने दिपि राजमार्ज मार्गमा साज छ छौदेखि बिहाला चार बजेस्सम्म सवारी चलाउन रो्ु। |
| small | ब्रतिक्यल मुस्मले उडान थब रुदा मन्त्ली मा तीं दिन देखी आलपत्र पर्यटकले सेना को जाहाज बात लुक्का पुर्यईने दिपी राज मरगमा साजज् चो देखी भीहाना चार बजे समस्वारी चलाओन रोग | रतिकुल मौसमले उडान ठप हुँदा मन्थलीमा तीन दिनदेखि अलपत्र पर्यटकलाई सेनाको जहाजबाट लुकला पुर्याइने विपि राजमार्गमा साइजत छ देखि बिहान चार बजे सम्बसवानी चलाउन रोग। |
| medium | प्रतिकुल मौसमले उडान थपपूदा मंथली मा तिन दिन देकि अलपत्र पर्यतक्रू सेना को जहाज बात लुकला पुर्याइनि भीपी राजमार्गमा साझ देकि भिहान चार बजे समसवारी चलाउन रोक | प्रतिकूल मौसमले उडान ठप हुँदा मन्थलीमा तीन दिनदेखि अलपत्र पर्यटकलाई सेनाको जहाजबाट लुकला पुर्याइने बीपी राजमार्गमा साझत 6 देखि बिहाल 4 बजेसम्म सवारी चलाउन रोग। |

## 5 Discussion and Conclusion

In this study, we focus on creating an exhaustive dataset and fine-tuning OpenAI's Whisper models for the task of Nepali language transcription, addressing key challenges such as dataset limitations and model overfitting. Our results demonstrate that dataset and vocabulary size, inclusivity, variability, model input compatibility, and augmentation strategies play a critical role in improving the model's transcription performance, as evaluated through WER.

The differences in performance between models trained on other open-source ASR datasets and our custom dataset are notable. The custom dataset provides the model with a more comprehensive representation of the linguistic variations in the Nepali language in terms of dialect, speaker accents, and environments compared to the Fleurs dataset used in Whisper's original paper and other open-source datasets compared here. This diversity allows the model to capture more nuanced features of the Nepali language, leading to better generalization. Comparing quantitatively, the fine-tuned models on the custom dataset show a significant improvement across all individual, combined, and augmented datasets. Combining the individual corpus and data augmentation further enhanced the model's performance. While we used relatively simple augmentation, it demonstrates that even minor augmentation techniques can significantly enhance transcription accuracy in low-resource languages like Nepali.

When comparing our fine-tuned models with OpenAI's Whisper models, the results show that our models significantly outperform the original Whisper models across all evaluated sizes—`tiny`, `base`, `small`, and `medium`. The progressive improvement across model sizes highlights the effectiveness of fine-tuning for domain-specific tasks, even with limited computing resources. While Whisper's training data, Fleurs consists of relatively short audio clips (2s-10s), our dataset contains longer and denser audio clips ranging from 5s to 30s. This wide range of clips is more compatible with Whisper's input, enabling the model to better capture contextual information over extended sequences.

In conclusion, our work shows the importance of dataset and their effectiveness on increasing the accuracy of speech-to-text model such as Whisper. Future work could explore more advanced augmentation techniques, fine-tuning larger Whisper models, and implementing similar fine-tuning approaches on other limited-resourced languages, potentially leading to further improvements in transcription accuracy.